\documentclass{article}

\usepackage{PRIMEarxiv}

\usepackage[utf8]{inputenc} 
\usepackage[T1]{fontenc}    
\usepackage{hyperref}       
\usepackage{url}            
\usepackage{booktabs}       
\usepackage{amsfonts}       
\usepackage{nicefrac}       
\usepackage{microtype}      
\usepackage{lipsum}
\usepackage{fancyhdr}       
\usepackage{graphicx}       
\graphicspath{{media/}}     
\usepackage{diagbox}
\usepackage{amsmath}

\title{Gaze Estimation Approach Using Deep Differential Residual Network} 

\author{
  \and
  Longzhao Huang\\
  School of Artificial Intelligence \\
  Guilin University of Electronic Technology \\
  Guilin 541004, China\\
  longzhaohuang1722@gmail.com \\
  \and
  Yujie Li*\\
  School of Artificial Intelligence \\
  Guilin University of Electronic Technology \\
  Guilin 541004, China\\
  yujieli@guet.edu.cn \\
  \and
  Xu Wang\\
  School of Artificial Intelligence \\
  Guilin University of Electronic Technology \\
  Guilin 541004, China\\
  1901610215@mails.guet.edu.cn \\
  \and
  Haoyu Wang\\
  School of Artificial Intelligence \\
  Guilin University of Electronic Technology \\
  Guilin 541004, China\\
  1901620216@mails.guet.edu.cn \\
  \and
  Haoyu Wang\\
  School of Artificial Intelligence \\
  Guilin University of Electronic Technology \\
  Guilin 541004, China\\
  1901620216@mails.guet.edu.cn \\
  \and
  Ahmed Bouridane \\
  Faculty of Engineering and Environment \\
  Northumbria University \\
  Newcastle NE18ST , UK \\
  abouridane@sharjah.ac.ae \\
  \and
  Ahmad Chaddad* \\
  The Laboratory for Imagery Vision and Artificial Intelligence \\
  Ecole de Technologie Superieure \\
  Montreal QC H3C1K3, Canada \\
  ahmad8chaddad@gmail.com \\
}

\begin{document}
\maketitle

\begin{abstract}
Gaze estimation, which is a method to determine where a person is looking at given the person's full face, is a valuable clue for understanding human intention. Similarly to other domains of computer vision, deep learning (DL) methods have gained recognition in the gaze estimation domain. However, there are still gaze calibration problems in the gaze estimation domain, thus preventing existing methods from further improving the performances. An effective solution is to directly predict the difference information of two human eyes, such as the differential network (Diff-Nn). However, this solution results in a loss of accuracy when using only one inference image. We propose a differential residual model (DRNet) combined with a new loss function to make use of the difference information of two eye images. We treat the difference information as auxiliary information. We assess the proposed model (DRNet) mainly using two public datasets (1) MpiiGaze and (2) Eyediap. Considering only the eye features, DRNet outperforms the state-of-the-art gaze estimation methods with $angular-error$ of 4.57 and 6.14 using MpiiGaze and Eyediap datasets, respectively. Furthermore, the experimental results also demonstrate that DRNet is extremely robust to noise images.
\end{abstract}


\section{Introduction}\label{sec1}

\begin{figure}[p]
\centering
\hspace{-90pt}
	\begin{minipage}{0.3\linewidth}
		\centering
		\includegraphics[width=0.8\linewidth]{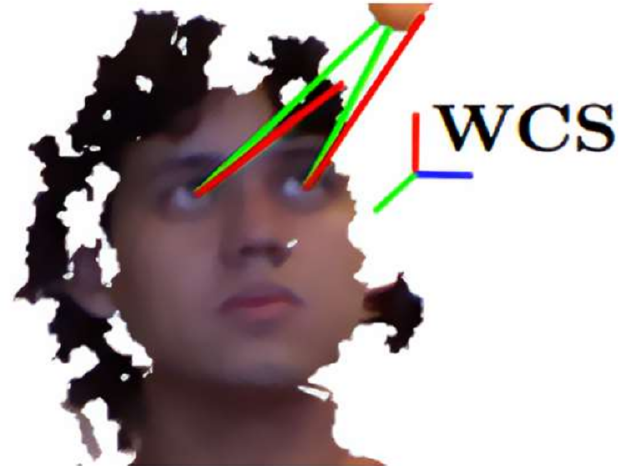}
	\\(\textbf{a})
	\end{minipage}
	\begin{minipage}{0.54\linewidth}
		\centering
 \includegraphics[width=0.9\linewidth]{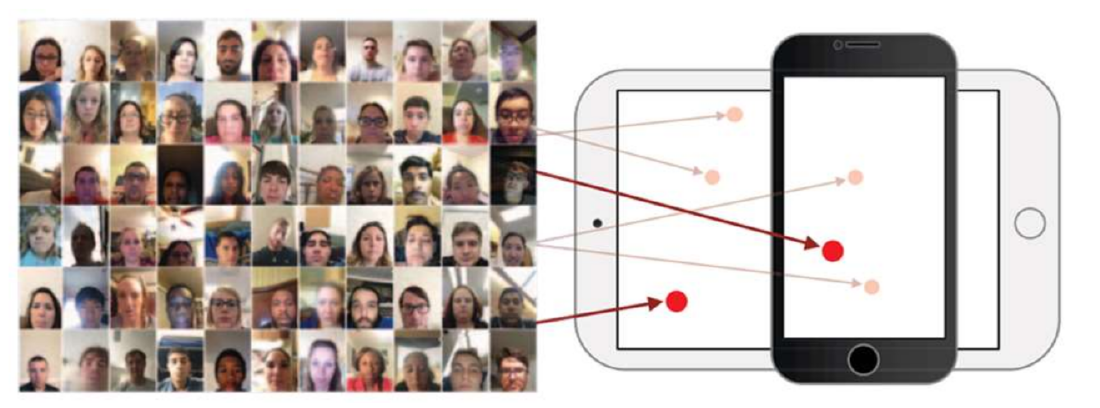}
 \\(\textbf{b})
	\end{minipage}\\
	\qquad
	\hspace{-120pt}
		\begin{minipage}{0.75\linewidth}
		\centering
 \includegraphics[width=1.02\linewidth]{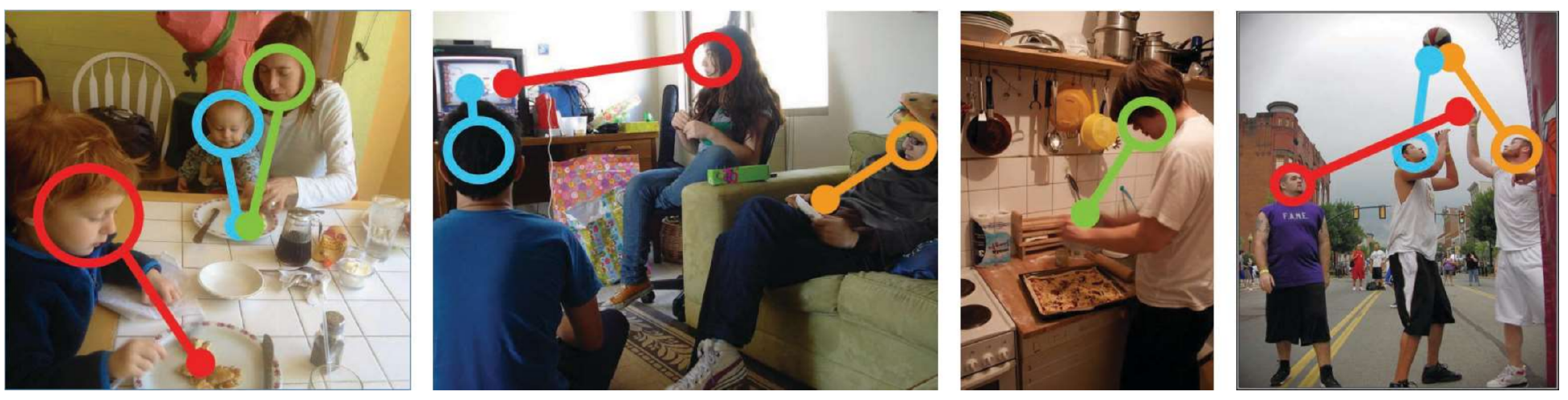}\\	(\textbf{c})
\end{minipage}

\caption{Examples of common tasks of gaze estimation: (\textbf{a}) three-dimensional (3D)-based estimation 
 \cite{funes2016gaze}, (\textbf{b}) target estimation \cite{2021Gaze, krafka2016eye} and (\textbf{c}) tracking \cite{recasens2015they}.}
 \label{fig1}
\end{figure}

Eye gaze is an important nonverbal communication technology. It contains rich information about human features, allowing researchers and users to tap more about human patterns \cite{eckstein2017beyond,li2018icassp} and action \cite{meissner2019promise,Haitham2019Inverse}. It is widely recommended in many topics, e.g., human--robot interaction (HRI) \cite{li2019appearance, 2022li,2010Controlling, 2022gazefordrive}. Most common gaze estimation tasks are categorized into three types: (1) three-dimensional (3D)-based gaze estimation \cite{funes2016gaze}, (2) target estimation \cite{2021Gaze,krafka2016eye} and (3) tracking estimation \cite{recasens2015they}. Figure \ref{fig1} shows examples of gaze estimation task types. However, our study focuses on 3D gaze estimation.

\begin{figure}[p]
	\begin{minipage}{0.44\linewidth}
		\centering
		\includegraphics[width=1\linewidth]{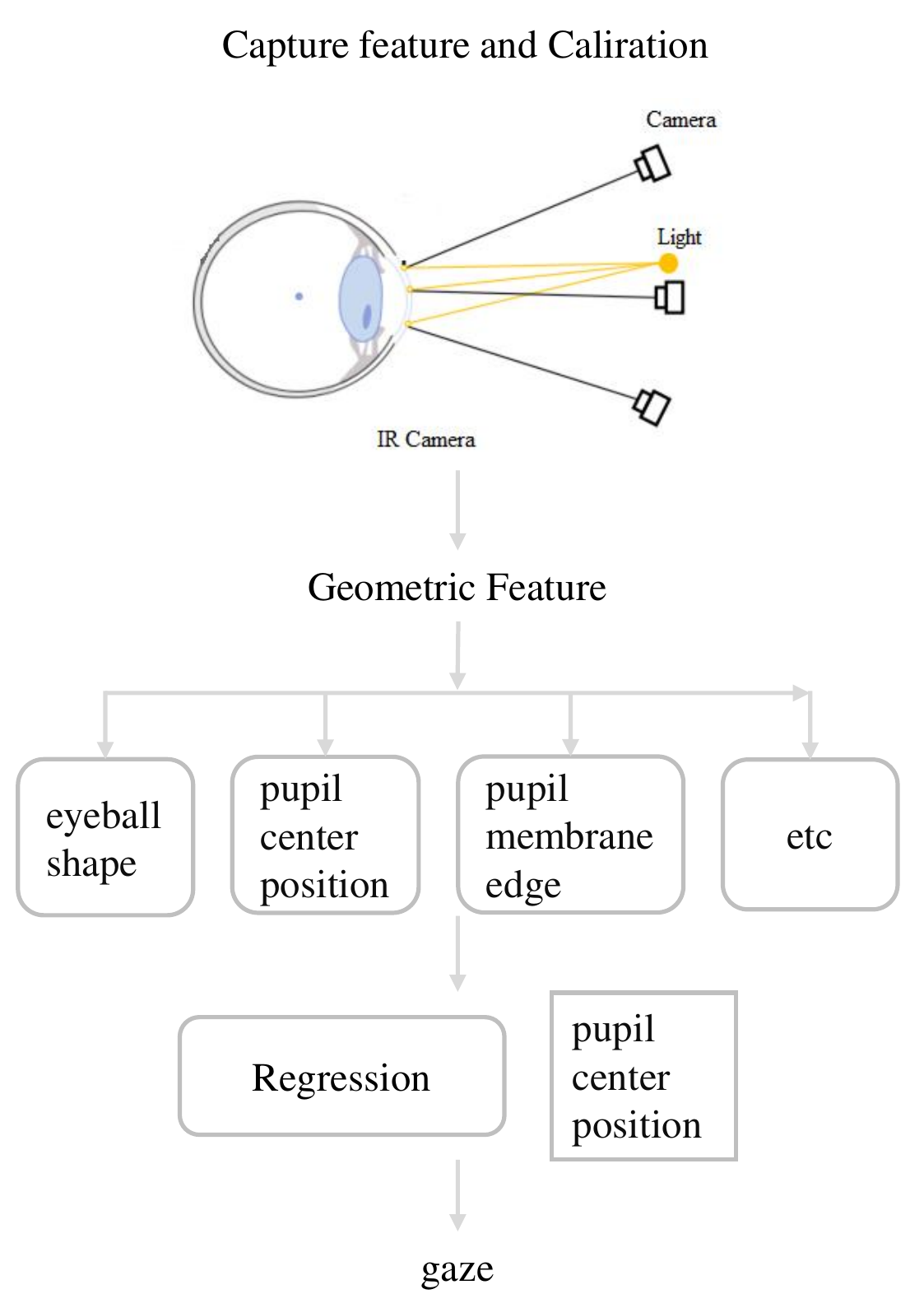}\\(\textbf{a})
	\end{minipage}
	\begin{minipage}{0.54\linewidth}
		\centering
 \includegraphics[width=1\linewidth]{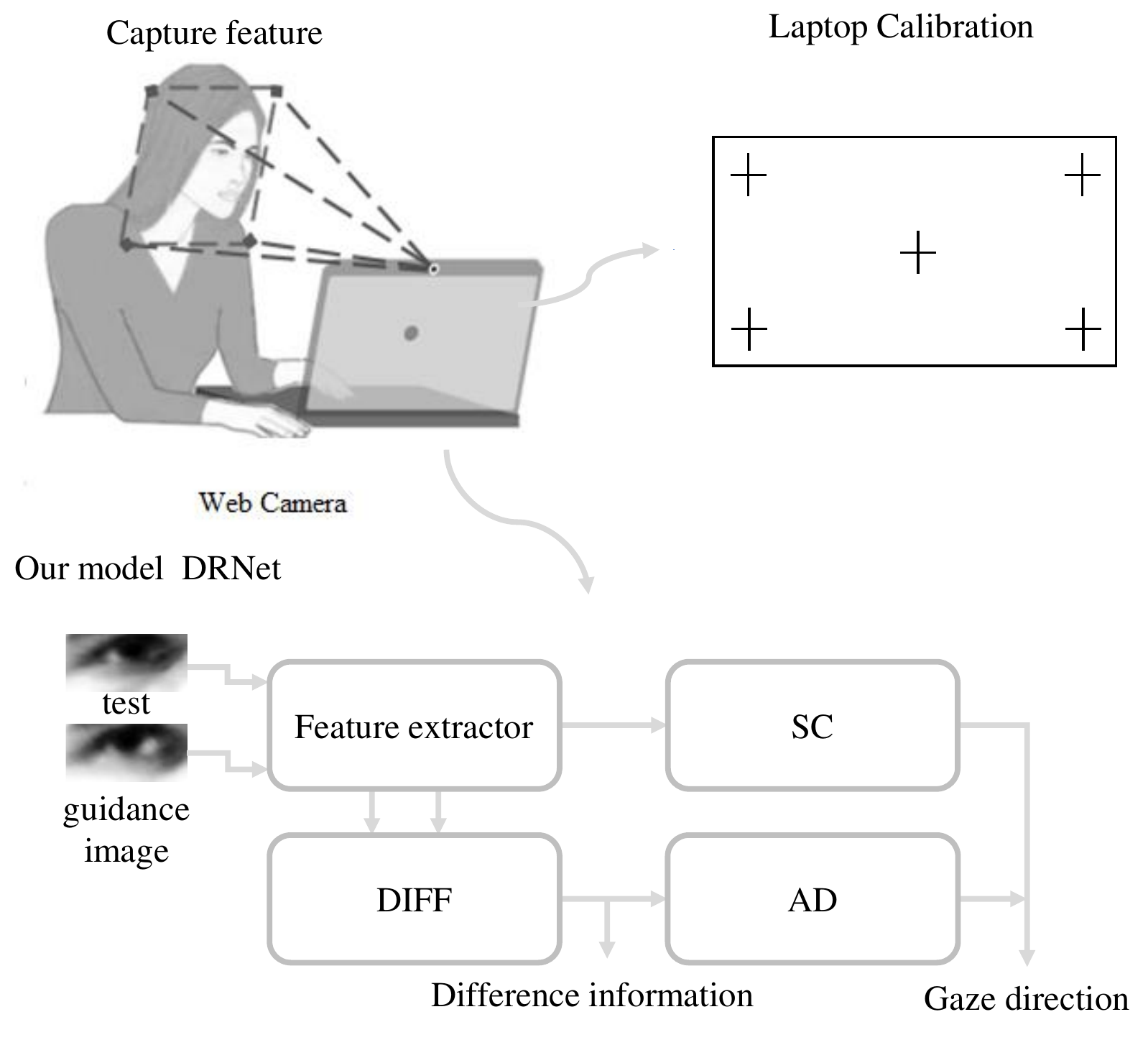}\\(\textbf{b}) 
	\end{minipage}

\caption{Example of two 3D gaze estimation techniques, (\textbf{a}) model-based and (\textbf{b}) appearance-based methods.}
 \label{fig2}
\end{figure}

Three-dimensional gaze estimation can be classified into two methods, as illustrated in (Figure \ref{fig2}). Model-based methods \cite{guestrin2006general,zhu2007novel,valenti2011combining,alberto2014geometric} generally consider geometric features such as eyeball shape, pupil center position, and pupil membrane edge. These methods require specific equipment such as infrared camera and have low robustness when illumination and head pose change. However, appearance-based methods have higher performance due to the training of a deep network using a large amount of data. Specifically, the deep network has the ability to extract features from eye images under various illumination conditions and head positions. Only a laptop with a web camera is required to collect the data set (e.g., MpiiGaze \cite{zhang2017mpiigaze}).

\unskip

\unskip

\begin{figure}[p]
	\centering
	\hspace{-94pt}
	\begin{minipage}{0.4\linewidth}
		\centering
		\includegraphics[width=0.85\linewidth]{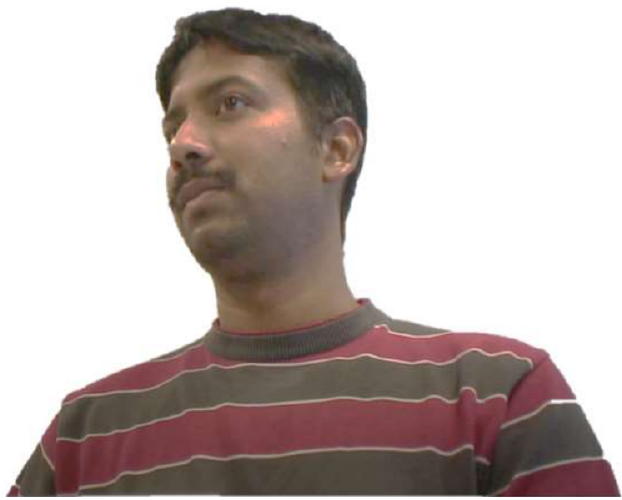}\\(a) 
	\end{minipage}
	\begin{minipage}{0.4\linewidth}
		\centering
 \includegraphics[width=0.9\linewidth]{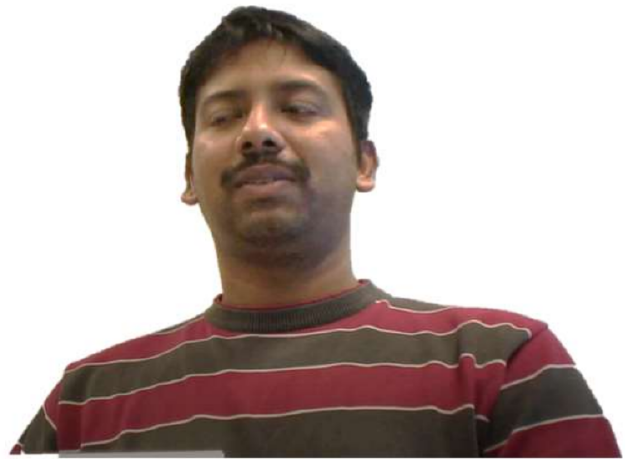}\\(b) 
	\end{minipage}\\
	\qquad
	\hspace{-114pt}
	\begin{minipage}{0.2\linewidth}
		\centering
 \includegraphics[width=0.7\linewidth]{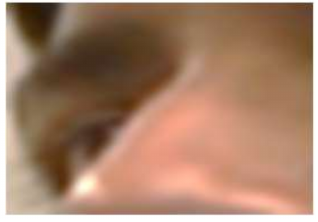}\\left eye in (a)
	\end{minipage}
		\begin{minipage}{0.2\linewidth}
		\centering
 \includegraphics[width=0.7\linewidth]{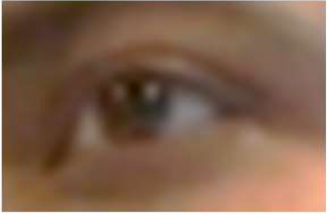}\\right eye in (a)
	\end{minipage}
	\begin{minipage}{0.2\linewidth}
		\centering
 \includegraphics[width=0.7\linewidth]{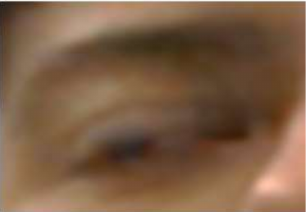}\\left eye in (b)
	\end{minipage}
	\begin{minipage}{0.2\linewidth}
		\centering
 \includegraphics[width=0.7\linewidth]{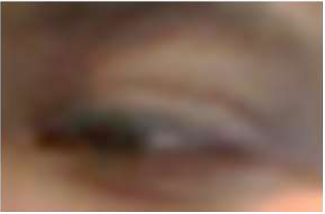}\\right eye in (b)
	\end{minipage}
\caption{Example
 of noise images in Eyediap
 \cite{funes2014eyediap}. (a) , (b) are two example frames of the dataset.} (\textbf{Top}) Original RGB frames. (\textbf{Bottom}) Left- and right-eye images are captured from the original RGB frames.
\label{fig3}
\end{figure}

The first appearance-based method \cite{zhang2015appearance} uses convolutional neural networks (CNN) inherited from LeNet \cite{lecun1998gradient} for gaze estimation. One factor that limits CNN success is the noise in the eye images. Figure \ref{fig3} shows the noise images caused by the extreme head position and the blink response in Eyediap \cite{funes2014eyediap}. The left eye image in (a) is not completely captured due to the extreme head position. The left and right eye images missed the pupil information in (b) due to the blink response. We aim to avoid the limitation of the noisy eye images using the proposed DRNet model (i.e., more details are given in Section \ref{sec4b}).

\unskip

Gaze calibration problem also limits the performances of CNNs. An effective and simple solution proposed by many publications to solve this problem is to adjust the weight of the model after training \cite{krafka2016eye,zhang2018training,li2020evaluation,linden2019learning,Wang2022}. However, this solution requires many inference images with the label. Liu et al. \cite{liu2019differential} propose a differential network (Diff-Nn) to address the gaze calibration problem by directly predicting the difference information between two images of the eyes. Gu et al. \cite{gu2021gaze} developed Diff-Nn for the gaze estimation using the left and right eye patch of one face simultaneously. Several other works mention that the performance based on the methods considering the difference information is directly affected by the number and the specific label of the inference image \cite{liu2019differential, gu2021gaze}.

We firstly treated the difference information as auxiliary information in the proposed DRNet. We combined the original gaze direction and the difference information through the shortcut-connection in DRNet. In addition, we proposed a new loss function for the gaze estimation. For example, the original loss function evaluates the gap between the quantity of the predicted vector and its ground truth, such as pitch and yaw. The new loss function evaluates the intersection angle between the predicted and its ground truth vector in {3D} space directly.

To the best of our knowledge, this is the first study that applies the shortcut-connection by combining the difference information to address gaze calibration. Our contributions can be summarized as follows.
\begin{itemize}
 \item We propose the DRNet model, which applies the shortcut connection, to address the gaze calibration problem and hence improve the robustness-to-noise image in the eye images. DRNet outperforms the state-of-the-art gaze estimation methods only using eye features, and is also highly competitive among the gaze estimation methods combining facial feature.
 \item We propose a new loss function for gaze estimation. It provides a certain boost to existing appearance-based methods.
\end{itemize}

The remainder of this paper is structured as follows. The related works are presented in Section \ref{sec2}. Section \ref{sec3} describes the proposed pipeline-based DRnet. We present the experimental results in Section \ref{sec4}. Finally, Section \ref{sec5} concludes the key contributions of our work.

\section{Related Work}\label{sec2}

In previous years, appearance-based methods have been considered as the most commonly methods in gaze estimation. For example, Zhang et al. \cite{zhang2015appearance} proposed the first appearance-base method (i.e., LeNet \cite{lecun1998gradient}) that uses eye features for gaze estimation. They expanded three convolution layers to sixteen convolution layers in their work \cite{zhang2017mpiigaze} to achieve higher performance metrics. Fischer et al. \cite{fischer2018rt} presented a two-stream network; left and right eye images are fed into VGG-16 \cite{simonyan2014very} separately. 
Some studies directly used face images as input or/and applied CNN to automatically extract deep facial features. For example, Zhang et al. \cite{zhang2017s} used a spatial weighting mechanism to efficiently encode the face location using CNN. This method decreases noise impact and improved the contribution of highly activated regions. Cheng et al. \cite{cheng2020coarse} assigned weights for two eye features under the guidance of facial features. Furthermore, Chen et al. \cite{chen2018appearance} considered dilated convolution to extract deep facial features. This effectively improves the perceptual field while reducing the image resolution \cite{chen2018appearance}. {In addition, gaze estimation in outdoor environments was investigated using eye and face features derived from near-infrared camera \cite{2018Deep,2020Deep}.} Bao et al. \cite{bao2021adaptive} studied a self-attention mechanism to combine two eye features with the guidance of facial features. In \cite{2022STF}, CNN with long short-term memory (LSTM) network is introduced to be able to capture spatial and temporal features from video frames. In \cite{2020GanforLLE}, the generative adversarial network is used to enhance the eye image captured under low and dark light conditions. Despite all the advantages of gaze estimation techniques, there are still some challenges that need to be addressed.

In order to avoid the challenges of previous gaze estimation techniques, we developed DRNet to treat the difference information as auxiliary information and designed the model based on the residual concept. It is worth noting that the residual network concept was first proposed by He et al. \cite{he2016identity} to avoid the model degradation problem of deep neural networks. For example, in residual networks, increasing the depth of the network does not result in decreasing the accuracy due to the shortcut connection. Thus, we apply the shortcut connection in DRNet to improve the robustness of the differential network. 

\section{Methodology}\label{sec3}
This paper proposes a DRNet model with a new loss function to optimize the performance of gaze estimation. Specifically, the difference information is used as an auxiliary information in DRNet model. A brief overview of the DRNet model with the proposed loss function are detailed as follows.

\subsection{Proposed DRNet}\label{sec3a}

\begin{figure}[p]
 \includegraphics[width=1\textwidth]{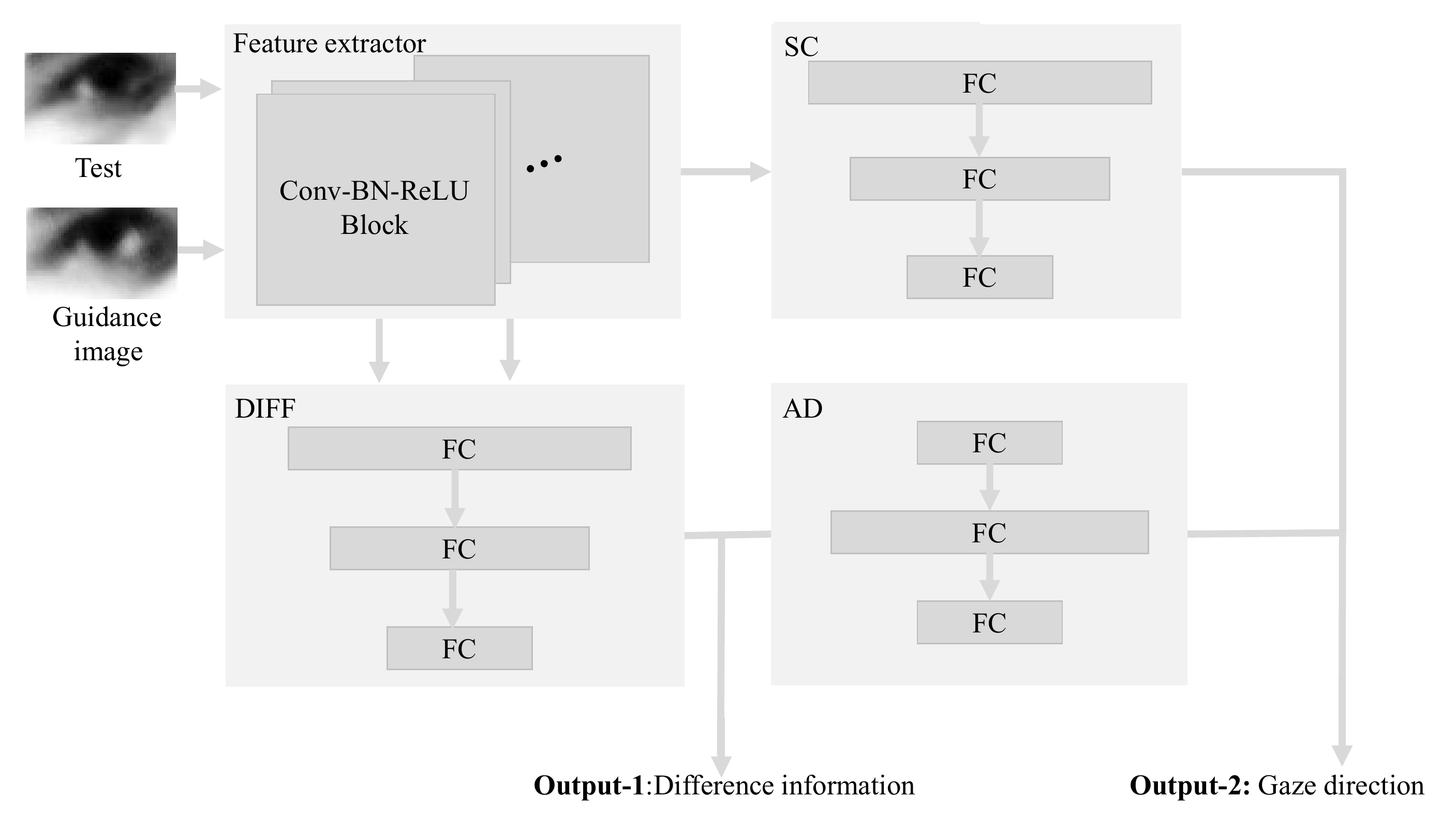}
\caption{The proposed DRNet pipeline. Test and guidance images used as input to the DRNet model. DRNet provides the difference information and the gaze direction, which are described by three-dimensional vectors.}
\label{fig4}
\end{figure}

Figure \ref{fig4} shows the proposed DRNet pipeline. It consists of a feature extractor, differential (DIFF), adjustment (AD), and shortcut (SC) modules. Specifically, DRnet receives two eye images (i.e., test and guidance images), and one of these eye images (i.e., guidance image) represents the calibration image. Furthermore, two eye input images are required to be derived from the same person.

\subsubsection{Feature Extractor}
Instead of one single eye image, both test and guidance eye images are adopted as raw input for DRNet. The feature extractor is stacked by the convolution layer (Conv), the batch normalization layer (BN) and the rectified linear unit (ReLU). The features are then used as derived from the fully connected layers.

\subsubsection{Residual Branch}
The three other components (i.e., DIFF, AD, and SC modules) construct the residual branch of the proposed DRNet architecture. More specifically, the DIFF module is responsible for providing the difference information between the test and guidance images. The AD module converts the difference information to the auxiliary information. The SC module provides the gaze-estimation-based information of the test image. Finally, the gaze direction represents the summation of SC and AD outputs.

\unskip

\subsection{The Residual Structure in DRNet}\label{sec3b}
\begin{figure}[p]
 \includegraphics[width=1\textwidth]{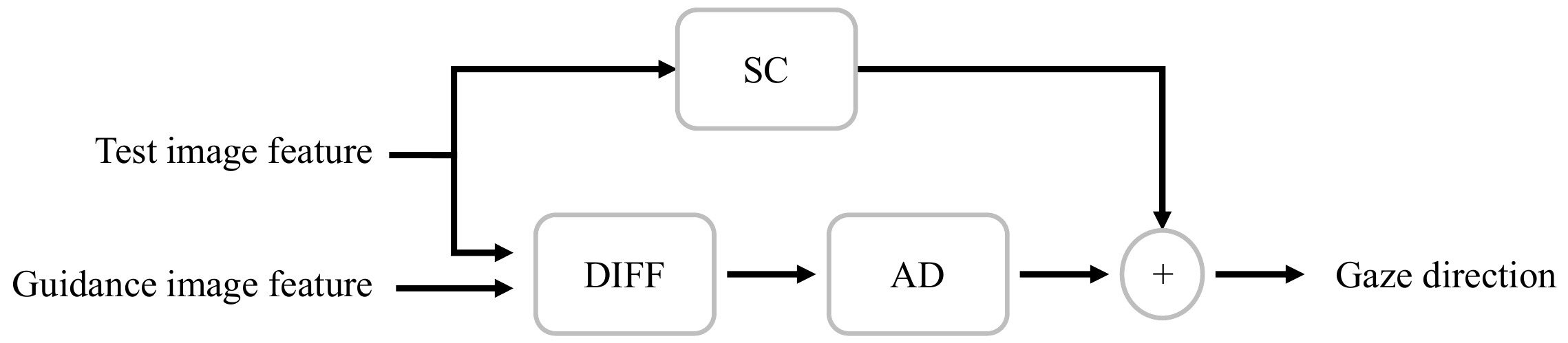}
\caption{DRNet residual structure. The guidance image and the test image features are extracted from the raw inputs. The sum of SC and AD outputs provides the gaze direction.}
\label{fig5}
\end{figure}

Figure \ref{fig5} shows a block diagram of residual structure process. Guidance and test image features are extracted by the feature extractor. These features are the input of the DIFF Module. In addition, test image feature is transferred into SC Module separately.

It is worth noting that the residual structure of our DRNet model is designed based on the ResNet architecture \cite{he2016identity}. Referring to the idea of a shortcut connection in ResNet \cite{he2016identity}, DRNet combined the difference information and gaze direction through the shortcut connection. The residual structure in DRNet is constructed by the fully-connected layer, while the residual structure in the ResNet is constructed based on the convolutional layers. Therefore, the residual structure of ResNet is an operation on the feature map, and the final output is the sum of two feature maps. Thus, the residual structure of DRNet operates on a one-dimensional vector, while the final output is the sum of two one-dimensional vectors.

\unskip

\subsection{Loss Function}\label{sec3c}

We propose a new and original loss function as follows:
\begin{equation}
L_{\text{new}} = {\frac{\lvert{g_{DRNet}\rvert \lvert{\hat g}_{test}\rvert}}
{\sqrt {g_{DRNet}} \sqrt {{\hat g}_{test}}}}, 
\end{equation}
\begin{equation}
L_{\text{original}} = \lvert g_{DRNet}-{\hat g_{test}}\rvert,
\end{equation}where $g_{DRNet}$ is the DRNet output, $g_{test}$ is the test image (e.g., the ground truth).

The loss functions $L_{\text{new}}$ and $L_{\text{original}}$ measure the angle and difference between the predicted vector and the ground truth vector, respectively. It is noted that the $L_{\text{new}}$ loss function uses approximate global information to optimize the output. Another loss function (named $LB$) based on a combined loss function $L_{\text{new}}$ and $L_{\text{original}}$. $LB$ can be expressed as follows:
\begin{equation}
LB{\text{ = }}\alpha {\text{*}}{\frac{\lvert{g_{DRNet}\rvert \lvert{\hat g_{test}}\rvert}}
{\sqrt {g_{DRNet}} \sqrt {{\hat g}_{test}}}} + (1 - \alpha )*\lvert g_{DRNet}-{\hat g_{test}}\rvert,
\end{equation} 
where $\alpha$ is the hyperparameter tuning $L_{\text{new}}$ and $L_{\text{original}}$.

We note that LB plays the role of optimization in DRNet.
We also optimized the DIFF module using the following loss function $LA$.
\begin{equation}
LA = {\lvert{\frac{{\lvert {{g_{diff}}} \rvert\lvert {{{\hat g}_{guidance}}} \rvert}}
{{\sqrt {{g_{diff}}} \sqrt {{{\hat g}_{guidance}}} }}{{ - }}\frac{{\lvert {{{\hat g}_{test}}} \rvert\lvert {{{\hat g}_{guidance}}} \rvert}}
{{\sqrt {{{\hat g}_{test}}} \sqrt {{{\hat g}_{guidance}}} }}} \rvert},
\end{equation}
where $g_{diff}$ and {$\hat g_{guidance}$} is the DIFF output and guidance image (i.e., ground-truth), respectively.
We use the loss function $LA$ to measure the difference information of the DIFF module.

 Compared to the loss function described in $Diff-Nn$ \cite{liu2019differential}, $LA$ also optimizes prediction by measuring difference information. In other words, $LA$ tunes the prediction to a reasonable scale. The process advantage of $LA$ is that the guidance image label will not be involved in the testing stage, while in $Diff-Nn$ some label information is needed.

The general loss function $L$ combined $LA$ and $LB$ as follows:
\begin{equation}
L = (1{{ - }}\beta )*LA + \beta *LB\ ,
\end{equation}
where $\beta$ is a hyperparameter tuning of $LA$ and $LB$. 

\subsection{Training Model}\label{sec3d}
\begin{figure}[p]
 \includegraphics[width=1\textwidth]{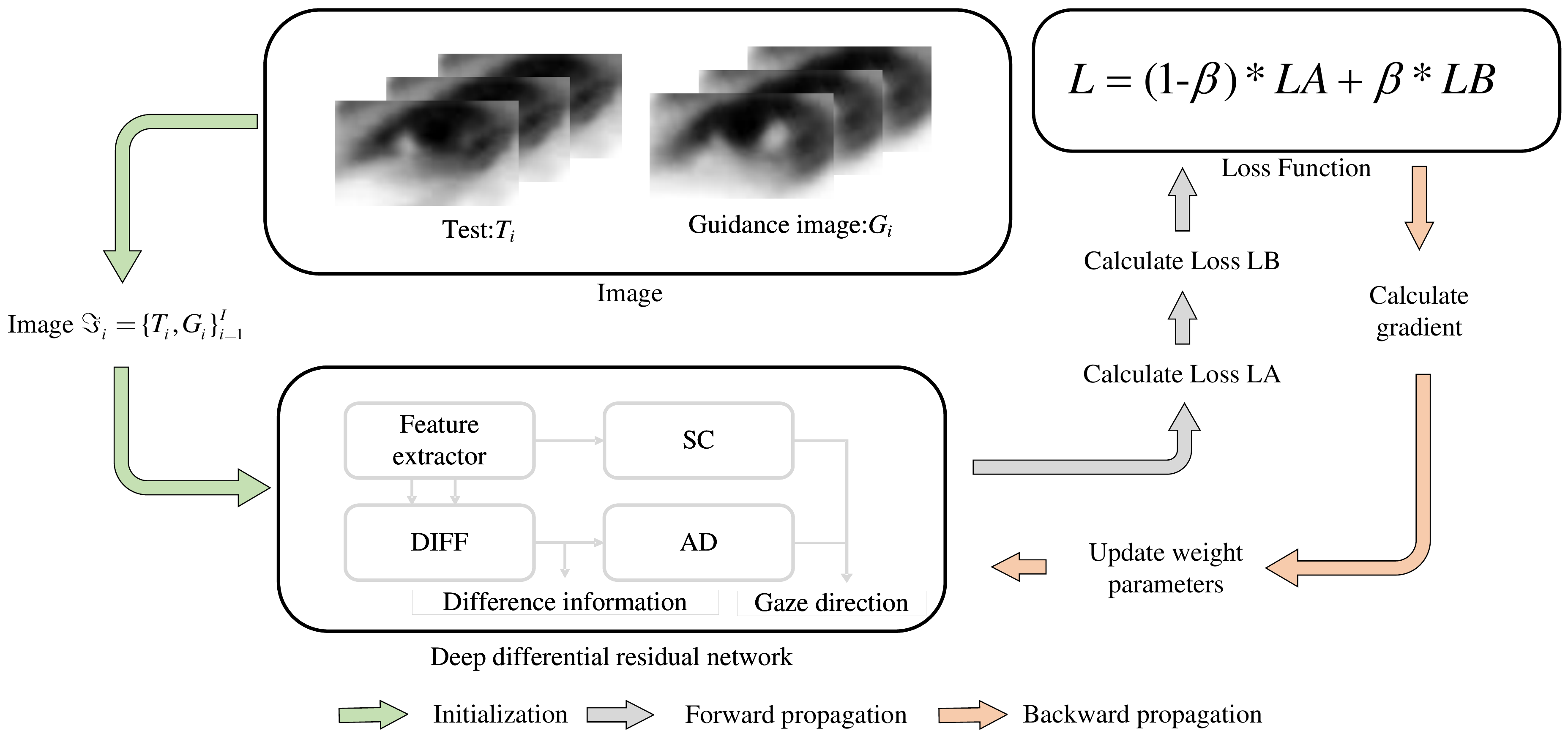}
 \caption{Flow chart
 of the training stage.}
\label{fig6}
\end{figure}

Figure \ref{fig6} shows the pipeline of the training model. (1) Initialization: Test image is fixed and guidance image randomly selected. (2) Forward propagation: Calculate the output of each unit, and the deviation between the target value and the actual output. (3) Backward propagation: compute the gradient and update the weight parameters. When the iteration reaches the maximum epoch, the DRNet parameters considered and fixed for the prediction.
We implement DRNet using PyTorch (\url{https://pytorch.org/}, accessed on February 2,  2020 
) that runs on TITAN RTX GPUs. We considered Adam optimizer with an initial learning rate of 0.01 (decayed by 0.1 every 5 epochs) and batch size of 128 and 1 in training and testing, respectively.

\unskip

\section{Experiments}\label{sec4}

\begin{figure}[p]
	\centering
		\hspace{-36pt}
	\begin{minipage}{0.48\linewidth}
		\centering
		\includegraphics[width=0.85\linewidth]{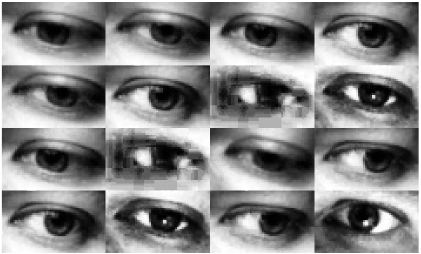}\\(\textbf{a}) 
		 \label{fig14}
	\end{minipage}
	\begin{minipage}{0.48\linewidth}
		\centering
 \includegraphics[width=0.85\linewidth]{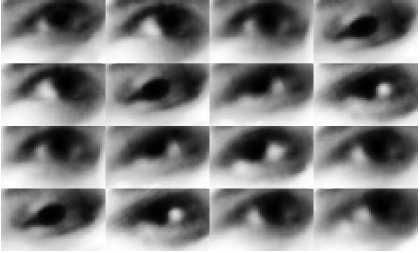}\\(\textbf{b}) 
 \label{fig15}
	\end{minipage}

\caption{Sample images from MpiiGaze (\textbf{a}) and Eyediap (\textbf{b}).} 
 \label{figDataset}
\end{figure}

To validate our proposed architecture, two public datasets have been used in the experimentation process: (i) MpiiGaze dataset and (ii) Eyediap dataset. An example of sample eye images in Eyediap and MpiiGaze is shown in Figure \ref{figDataset}.

\unskip

(1) MpiiGaze dataset consists of 1500 left and right eye images derived from \mbox{15 subjects \cite{zhang2017mpiigaze}}. These images are obtained in real-life scenarios with a variation of the illumination conditions, head pose and subjects with glasses. Specifically, the images are grayscale with a resolution of $36 \times 60$ pixels with corresponding information related to head pose. (2) Eyediap dataset consists of 94 videos taken from 16 subjects \cite{funes2014eyediap}. The data set is obtained in a laboratory setting with the corresponding head pose and gaze. Data sets were pre-processed following the pre-processing procedures described in \cite{cheng2021appearance}, and cropping approximately 21K images of the eyes, which are also grayscale images with a size of \mbox{36 × 60 pixels}. Note that since two subjects lack the videos in the screen target session, we obtained the images of 14 subjects in our experiments. 

We used $angular-error$ as a measurement which is generally used to measure the accuracy of the 3D gaze target method as follows:
\begin{equation}
angle\_error = \frac{{\lvert {{g_{test}}} \rvert\lvert {{{\hat g}_{test}}} \rvert}}
{{\sqrt {{g_{test}}} \sqrt {{{\hat g}_{test}}} }}.
\end{equation}
where ${\hat g}_{test}$ and $g_{test}$ is the true and predicted test image for gaze direction, respectively.



\subsection{Appearance-Base Methods}\label{sec4a}

Table \ref{tab1} reports the $angular-error$ of the appearance-based methods using eye features. Compared to baseline methods (Mnist \cite{zhang2015appearance}, GazeNet \cite{zhang2017mpiigaze}, RT-Gen \cite{fischer2018rt}, DenseNet101-Diff-Nn \cite{liu2019differential}), our proposed DenseNet101-DRNet delivers the best performance using the features of the eye image of two public datasets (Figure \ref{figDataset}). It is worth noting that the loss function used in DRNet is based on Equation (5), where $\alpha$ = $\beta$ = 0.75. 

\begin{figure}[p]
 \centering
 \begin{minipage}{0.49\linewidth}
 \centering
 \includegraphics[width=1\hsize]{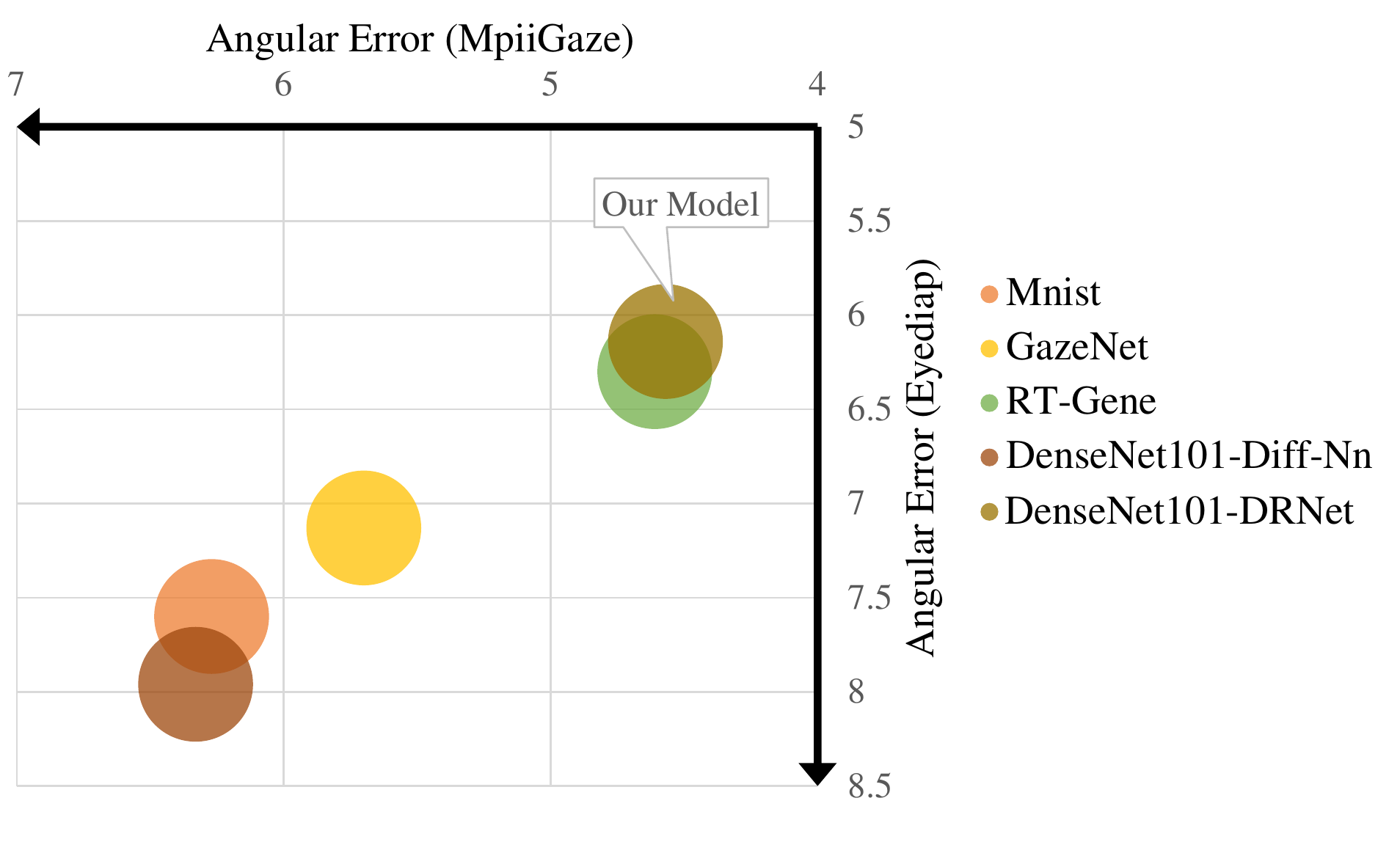}\\(\textbf{a})
 \label{fig7}
 \end{minipage} 
 \begin{minipage}{0.49\linewidth}
 \centering
 \includegraphics[width=1\hsize]{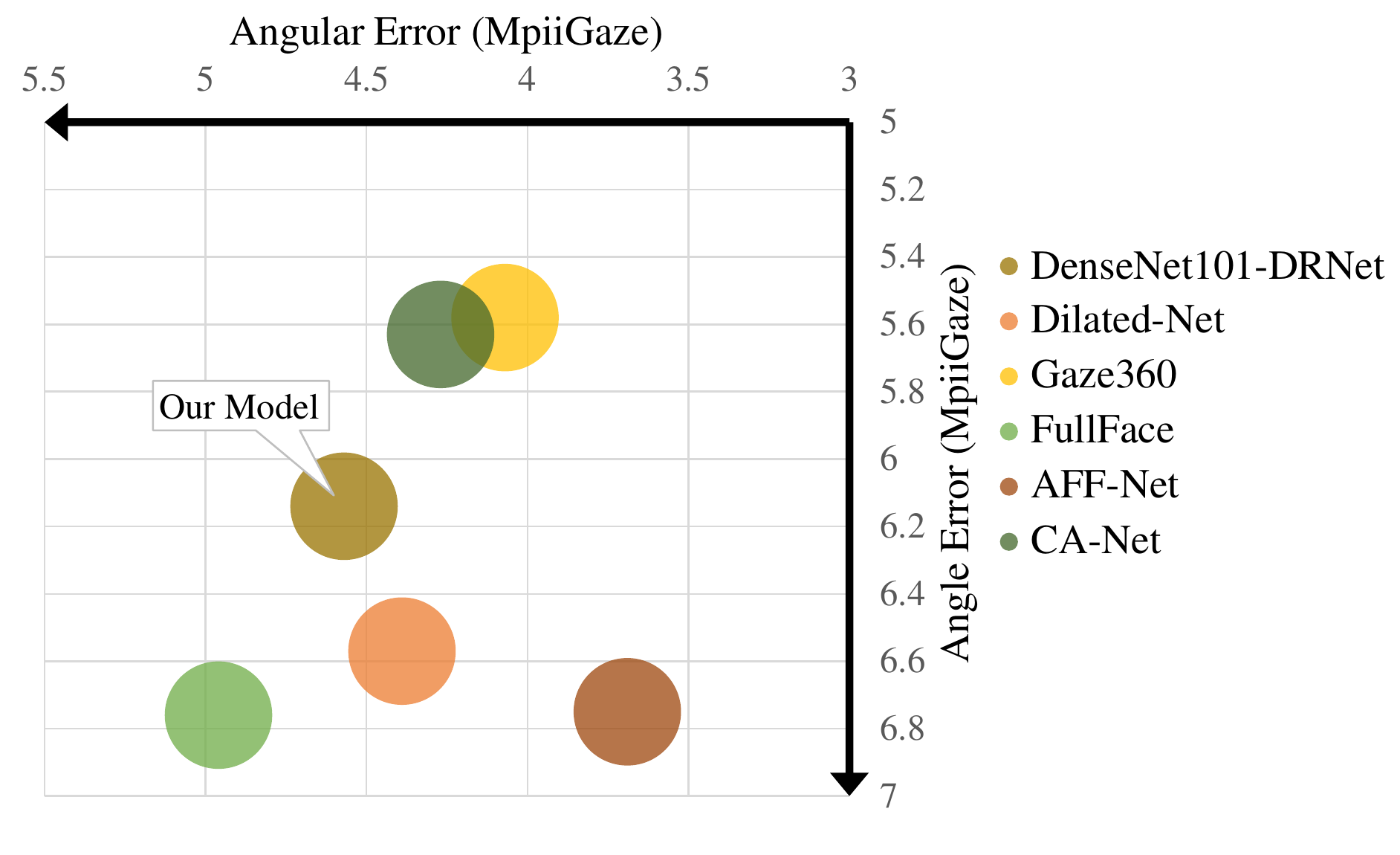}\\(\textbf{b})
 \label{fig8}
 \end{minipage} 
\caption{Performance ($angular-error$) of methods that use (\textbf{a}) eye features or (\textbf{b}) combined eye and facial features. The horizontal axis records the error in the MpiiGaze dataset and the vertical axis records the error in the Eyediap dataset. The model closest to the upper right corner represents better performance.}
\label{fig7&8}
\end{figure}

{ We also compared the performance of the proposed DRNet with the baseline methods using eye and facial features (i.e., Dilated-Net \cite{chen2018appearance}, Full Face \cite{zhang2017s}, Gaze360 \cite{kellnhofer2019gaze360}, AFF-Net \cite{bao2021adaptive}, CA-Net \cite{cheng2020coarse}). We note that DRNet model uses only eye features. In related works, the performance of methods using eye and facial features show higher performance than the methods using only the eye features. Table \ref{tab2} reports the performance of $angular-error$. It can be seen that the proposed DenseNet101-DRNet is highly competitive among the gaze estimation methods (Figure \ref{fig7&8}). DRNet model is better than Dilated-Net \cite{chen2018appearance}, FullFace \cite{zhang2017s}, AFF-Net \cite{kellnhofer2019gaze360} using Eyediap dataset. In addition, DRNet shows a better performance than FullFace \cite{zhang2017s} using MpiiGaze dataset.}

\begin{table}
 \caption{\textls[-25]{Performance summary ($angular-error$) and comparison with recent works using eye features.}}\label{tab1}
  \centering
  \begin{tabular}{lll}
    \toprule
    \textbf{Method} & \textbf{MpiiGaze} & \textbf{Eyediap}\\ 
    \midrule
    Mnist \cite{zhang2015appearance} & 6.27 & 7.6\\
    GazeNet \cite{zhang2017mpiigaze} & 5.7 & 7.13\\
    RT-Gene \cite{fischer2018rt} & 4.61 & 6.3 \\
    DenseNet101-Diff-Nn \cite{liu2019differential} & 6.33 & 7.96\\
    DenseNet101-DRNet (ours) & \textbf{4.57} & \textbf{6.14}\\
    \bottomrule
  \end{tabular}
\end{table}

\begin{table}
 \caption{Performance summary ($angular-error$) and comparison with recent works using eye and facial features.}\label{tab2}
  \centering
  \begin{tabular}{lll}
    \toprule
    \textbf{Method} & \textbf{MpiiGaze} & \textbf{Eyediap} \\ \midrule
    Dilated-Net \cite{chen2018appearance} & 4.39 & 6.57 \\
    Gaze360 \cite{kellnhofer2019gaze360} & 4.07 & \textbf{5.58} \\
    FullFace \cite{zhang2017s} & 4.96 & 6.76 \\
    AFF-Net \cite{bao2021adaptive} & \textbf{3.69} & 6.75 \\
    CA-Net \cite{cheng2020coarse} & 4.27 & 5.63 \\
    \bottomrule
  \end{tabular}
\end{table}

{ In addition, we assess DenseNet101-DRNet using the Columbia gaze dataset (CAVE-DB) \cite{Columbia}. We found that the DenseNet101-DRNet using CAVE-DB shows the lowest $angular-error$ of 3.70 compared to 4.57 and 6.14 using MpiiGaze and Eyediap datasets, respectively.} 

\unskip

\subsection{Noise Impact on DRNet Model}\label{sec4b}

\begin{figure}[p]
 \includegraphics[width=1\textwidth]{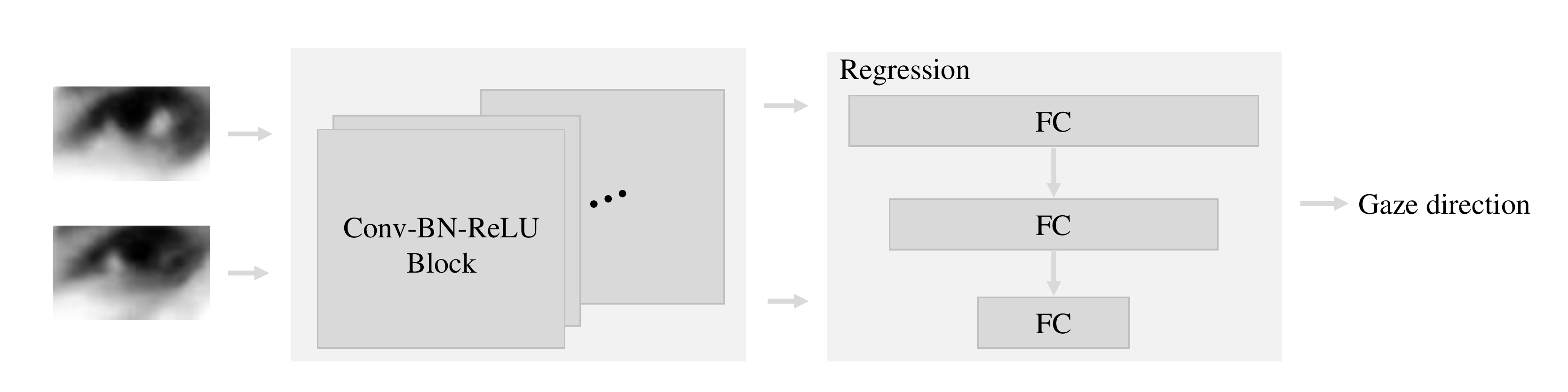}
\caption{Example of a two-stream model. The feature extractor module is based on convolution layers. Regression module is represented by fully connected layers.}
\label{fig11}
\end{figure}

To study the impact of noise impact on the proposed DRNet architecture, we have adopted RT-Gene (RT-Gene \cite{fischer2018rt} is a model using two eye images where the left and right eye patches are fed separately to VGG-16 networks \cite{simonyan2014very} allowing us to perform feature extraction) as a two-stream model. This scenario is used with an input using two images of the eye \cite{fischer2018rt}. Figure \ref{fig11} shows an example of a two-stream model. It consists of a feature extractor (e.g., convolution layers) and a regression (e.g., fully connected layers) modules. Two eyes images (i.e., test and guidance images) are used as raw input. The resulting output is a one-dimension vector that represents the gaze direction.
Likewise, the loss function used in DRNet is based on Equation (5), where $\alpha$ = $\beta$ = 0.75. While, the loss function based on Equation (3) used in the two-stream model, where $\alpha$ = 0.75.

\unskip

Again, we trained the two models using the two public datasets (i.e., MpiiGaze and Eyediap). In the validation step, the noise image was set as the guidance image. 

\begin{figure}[p]
 \centering
 \begin{minipage}{0.49\linewidth}
 \centering
 \includegraphics[width=1\hsize]{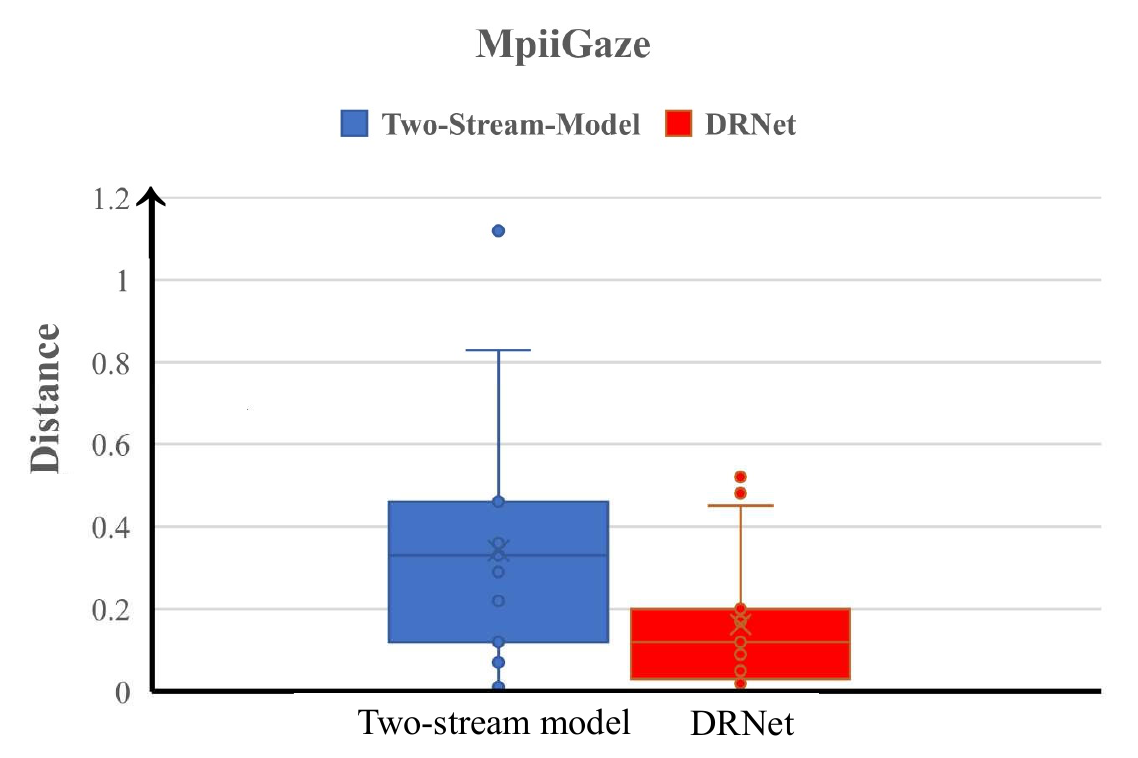}\\(\textbf{a})
 \label{fig9}
 \end{minipage} 
 \begin{minipage}{0.49\linewidth}
 \centering
 \includegraphics[width=1\hsize]{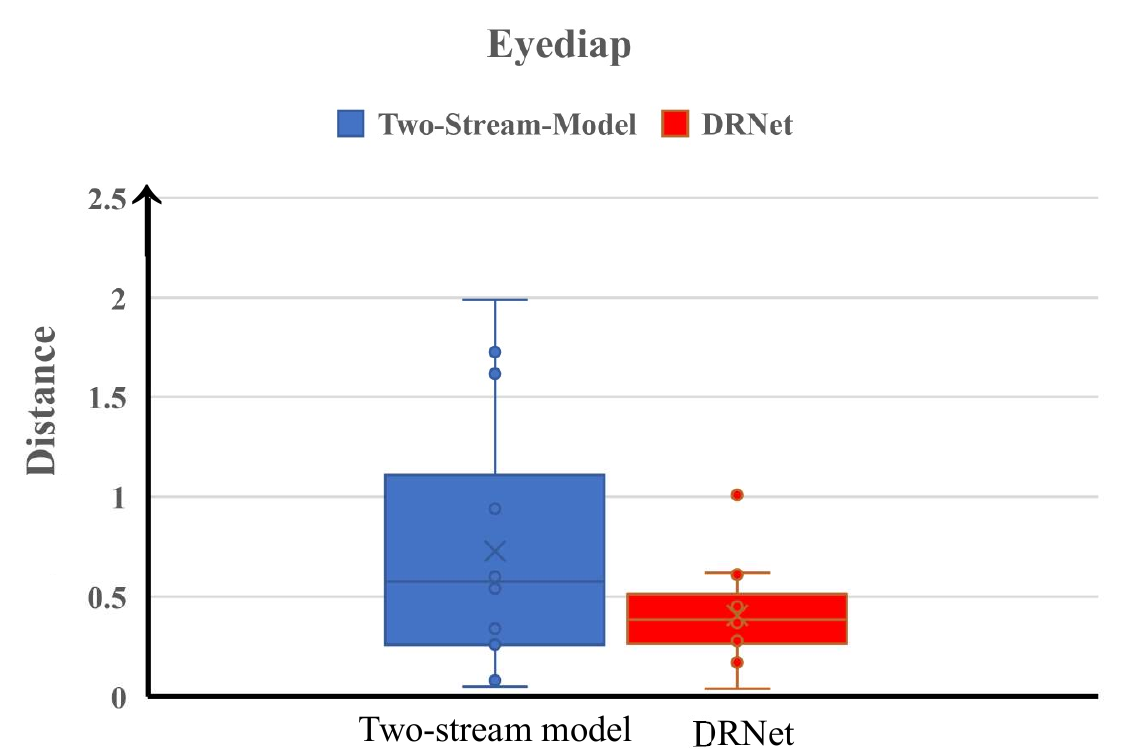}\\(\textbf{b})
 \label{fig10}
 \end{minipage} 
\caption{Box plots of the absolute distance of $angular-error$ between the normal and noisy image using the two-stream model and DRNet in MpiiGaze (\textbf{a}) and Eyediap (\textbf{b}).}
\label{fig9&10}
\end{figure}

Table \ref{tab3} reports the performance metrics ($angular-error$ and the absolute distance) using the two$-$stream model and DRNet, respectively. Additionally, we computed the absolute distance of difference $angular-error$ for each person between the normal and noisy image. When the distance is larger, the influence of the noise image is greater. It was observed that the average $angular-error$ and the absolute distance of the DRNet architecture were observed to be lower than the two$-$stream model using the MpiiGaze (i.e., normal image: two$-$stream model versus (vs) DRNet = 6.18 vs. 5.98; noisy image: two$-$stream model vs. DRNet = 6.39 vs. 5.99; distance (two$-$stream model vs. DRNet) = $0.34-0.16$) and Eyediap (i.e., normal image: two$-$stream model vs. DRNet = 7.07 vs. 6.71; noisy image: two$-$stream model vs. DRNet = 7.58 vs. 6.96; distance (two$-$stream model vs. DRNet) = $0.73-0.41$) datasets. Figure \ref{fig9&10} shows the distance metrics using the box plot function. From the results, we conclude that the DRNet architecture provides higher performance, as shown by a lower influence of the noise image compared to the two$-$stream model.

\unskip

\subsection{Assessing the Impact of the Loss Functions}\label{sec4c}
We conducted an experiment based on the Mnist network \cite{zhang2017mpiigaze} using the loss function of LB (Equation (3)) with the MpiiGaze and Eyediap datasets. Specifically, the Mnist model uses the original loss function (Equation (2)) where $\alpha$ = 0 and the new loss function where $\alpha=1$.

Table \ref{tab5} reports the $angular-error$ of the Mnist model. From the results, it was found that the MNIST model achieved the best performance with 7.27 in Eyediap and 6.07 in MpiiGaze when $\alpha$ in the range of $[0.75,1]$. We have also observed that the loss function LB provides much more optimized performance metrics.

\begin{table}
 \caption{Summary of $angular-error$ and absolute distance for robustness evaluation in the two-stream and DRNet model}\label{tab3}
  \centering
  \begin{tabular}{l|lll}
    \toprule
\multicolumn{4}{c}{\textbf{Two Stream Model versus DRNet}} \\ \hline
\textbf{MpiiGaze 
} & No\_Invalid\_Image & Fixed\_Invalid\_Image & Distance \\ \hline
$P_{00}$ & 4.72-
4.46 & 5.09-4.48 & 0.37-0.02 \\
$p_{01}$ & 5.98-5.99 & 6.11-6.02 & 0.13-0.03 \\
$p_{02}$ & 5.29-5.02 & 5.41-4.85 & 0.12-0.17 \\
$p_{03}$ & 6.65-6.61 & 6.43-6.41 & 0.22-0.20 \\
$p_{04}$ & 6.78-6.70 & 6.79-6.18 & 0.01-0.52 \\
$p_{05}$ & 6.20-6.27 & 7.32-6.39 & 1.12-0.12 \\
$p_{06}$ & 6.15-5.94 & 6.61-5.94 & 0.46-0.00 \\
$p_{07}$ & 7.44-7.19 & 7.11-7.07 & 0.33-0.12 \\
$p_{08}$ & 6.51-6.46 & 6.84-6.41 & 0.33-0.05 \\
$p_{09}$ & 7.98-7.07 & 7.91-7.09 & 0.07-0.02 \\
$p_{10}$ & 5.41-5.38 & 5.88-5.33 & 0.47-0.05 \\
$p_{11}$ & 5.26-4.88 & 5.26-5.36 & 0.00-0.48 \\
$p_{12}$ & 5.87-5.33 & 6.70-5.78 & 0.83-0.45 \\
$p_{13}$ & 6.39-6.11 & 6.68-6.20 & 0.29-0.09 \\
$p_{14}$ & 6.01-6.22 & 5.65-6.34 & 0.36-0.12 \\ \hline
Average & 6.18-5.98 & 6.39-5.99 & 0.34-\textbf{0.16} \\ \hline
\textbf{Eyediap}& No\_Invalid\_Image & Fixed\_Invalid\_Image & Distance \\ \hline
$p_{1}$ & 7.35-6.86 & 7.27-7.14 & 0.08-0.28 \\
$p_{2}$ & 7.43-7.33 & 6.87-7.66 & 0.56-0.33 \\
$p_{3}$ & 5.78-6.01 & 6.41-6.05 & 0.63-0.04 \\
$p_{4}$ & 7.66-5.29 & 7.92-5.58 & 0.26-0.29 \\
$p_{5}$ & 8.08-6.06 & 8.67-7.07 & 0.59-1.01 \\
$p_{6}$ & 7.14-5.84 & 7.40-6.21 & 0.26-0.37 \\
$p_{7}$ & 7.64-6.96 & 9.63-7.58 & 1.99-0.62 \\
$p_{8}$ & 8.23-5.44 & 9.17-5.61 & 0.94-0.17 \\
$p_{9}$ & 8.10-7.37 & 7.56-7.77 & 0.54-0.40 \\
$p_{10}$ & 7.24-7.87 & 8.86-8.32 & 1.62-0.45 \\
$p_{11}$ & 6.46-6.93 & 6.12-7.54 & 0.34-0.61 \\
$p_{14}$ & 5.35-7.78 & 5.30-7.56 & 0.05-0.22 \\
$p_{15}$ & 6.11-7.26 & 7.84-6.78 & 1.73-0.48 \\
$p_{16}$ & 6.46-7.00 & 7.06-6.58 & 0.60-0.42 \\ \hline
Average & 7.07-6.71 & 7.58-6.96 & 0.73-\textbf{0.41}\\ \noalign{\hrule height 1.0pt}
\end{tabular} 

\noindent{\footnotesize{No\_Invalid\_Image and Fixed\_Invalid\_Image represent the normal and noisy image, respectively. Distance represents the absolute value of difference $angular-error$ for each person between No\_Invalid\_Image and Fixed\_Invalid\_Image. (-): versus.}}

\end{table}

\begin{table}
 \caption{The performance of Mnist~\cite{li2019appearance} with different $\alpha$.}\label{tab5}
  \centering
  \begin{tabular}{lll}
    \toprule
     \boldmath{$\alpha$} & \textbf{Eyediap} & \textbf{MpiiGaze} \\
     \midrule
      1 & 7.31 & 6.07 \\
      0.75 & 7.27 & 6.12 \\
      0.5 & 7.38 & 6.25 \\
      0.25 & 7.59 & 6.53 \\ 
      0 & 7.6 & 6.3 \\ 
    \bottomrule
  \end{tabular}
\end{table}

\begin{figure}
 \centering
 \begin{minipage}{0.49\linewidth}
 \centering
 \includegraphics[width=1\hsize]{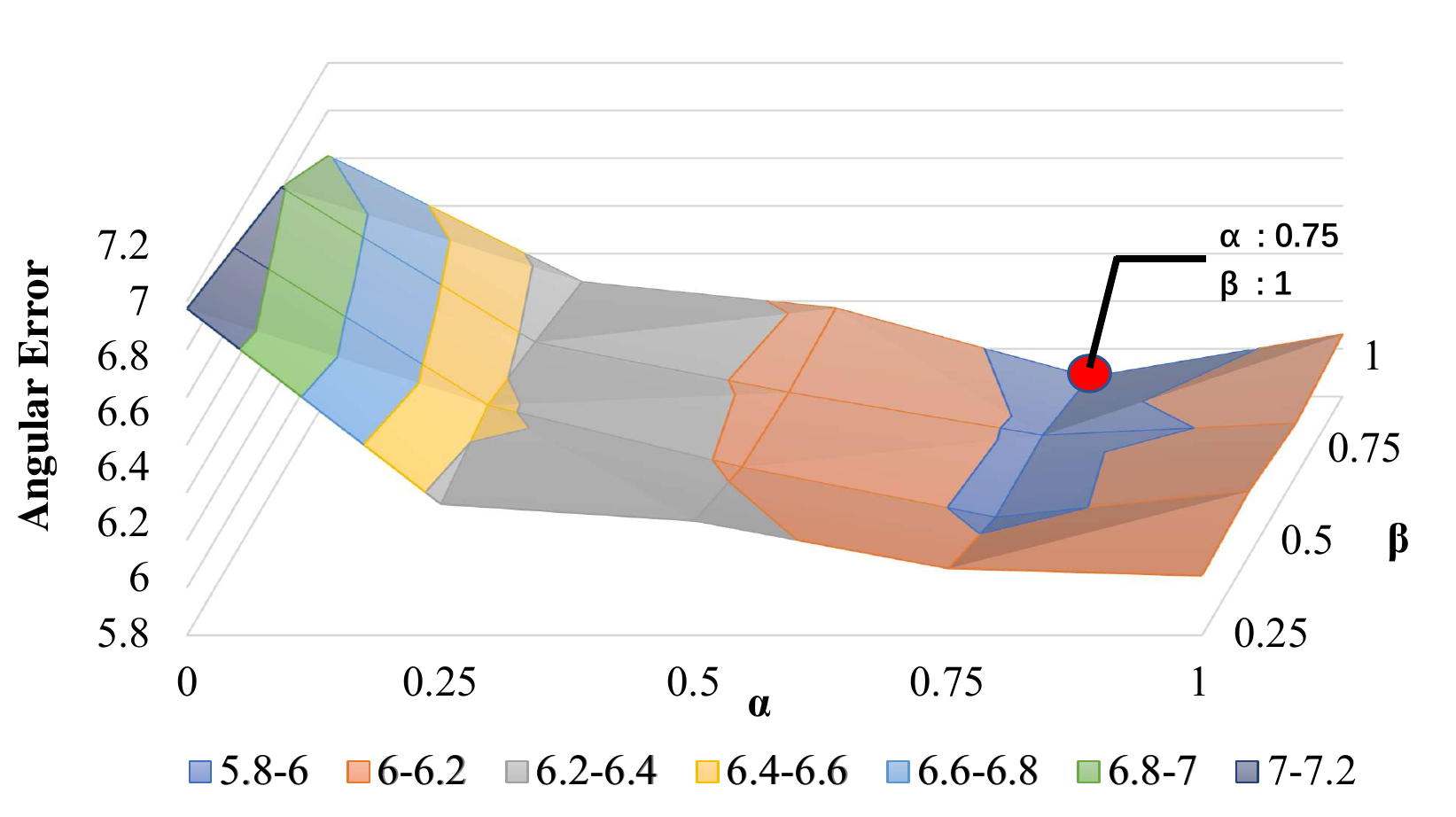}\\(\textbf{a})
 \label{fig12}
 \end{minipage} 
 \begin{minipage}{0.49\linewidth}
 \centering
 \includegraphics[width=1\hsize]{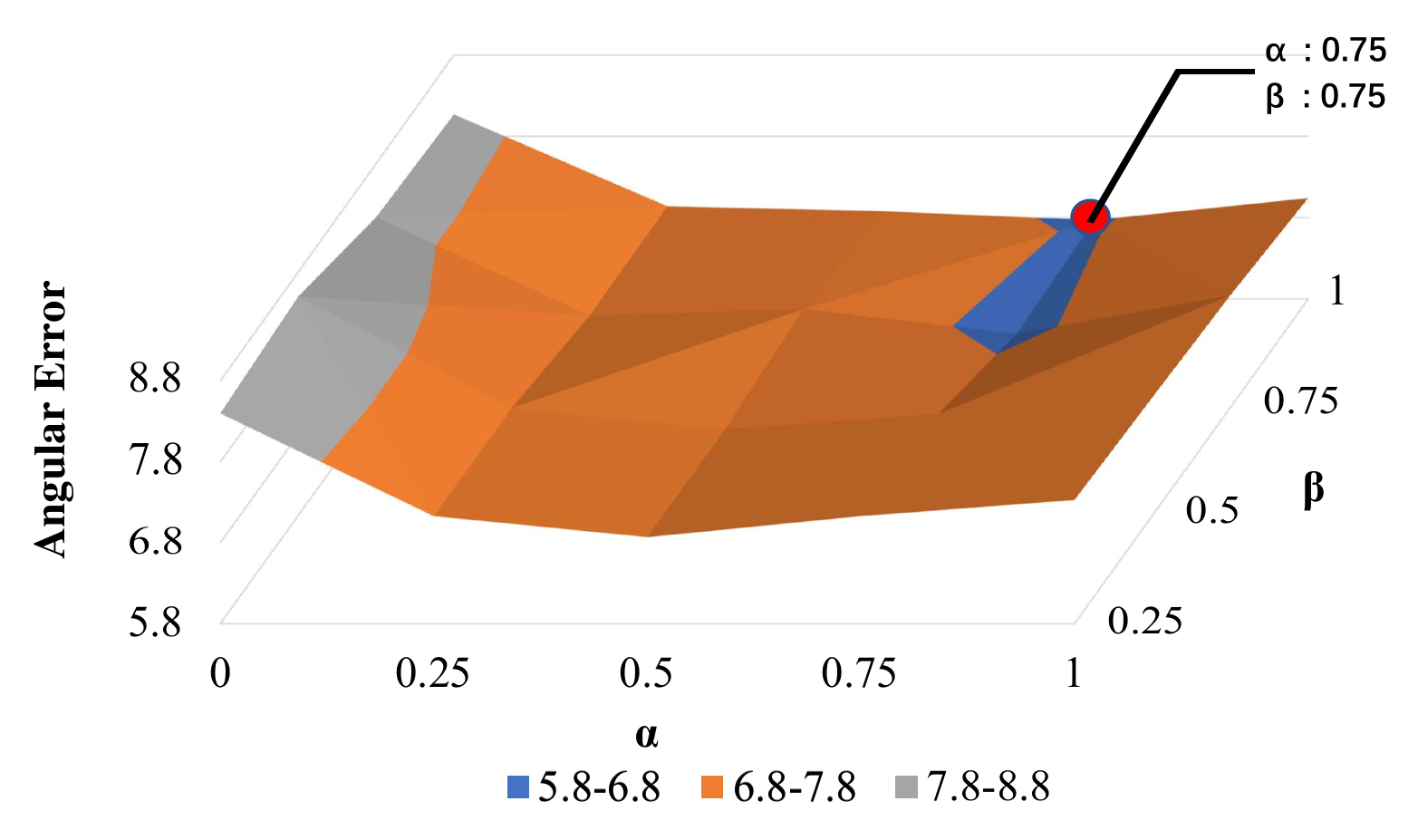}\\(\textbf{b})
 \label{fig13}
 \end{minipage}
\caption{Surface
plots the $angular-error$ 
in function with $\alpha$ and $\beta$ using MpiiGaze (\textbf{a}) and Eyediap (\textbf{b}). The red dot represents the best performance.}
\label{fig12&13}
\end{figure}

We also studied the impact of $\alpha$ and $\beta$ in DRNet architecture. The loss function used in DRNet with LA and LB is shown in Equation (5), Equation (4) and Equation (3), respectively. We set $\beta$ and $\alpha$ to 0.25, 0.5, 0.75, 1. Table \ref{tab6} reports the $angular-error$ of DRNet in function of $\beta$ and $\alpha$. We found the best $angular-error$ of 5.88 and 6.71 achieved when $\alpha$=0.75 and $\beta$ in the range of [0.75, 1] using the MpiiGaze and Eyediap datasets, respectively. Figure \ref{fig12&13} illustrates the surface of $angular-error$ as a function with $\beta$ and $\alpha$. As a trade-off, we set the hyperparameters to 0.75 for both $\alpha$ and $\beta$. 

\begin{table}
 \caption{The performance of DRNet in MpiiGaze/Eyediap with different $\alpha$ and $\beta$.}\label{tab6}
  \centering
  \begin{tabular}{l|cccc}
    \noalign{\hrule height 1.0pt}
    \diagbox{\boldmath{$\alpha$}}{\boldmath{$\beta$}} & 0.25 
    & 0.5 & 0.75 & 1 \\ \hline
    0 & 7.17/8.4 & 7.09/8,5 & 7.01/8.14 & 6.81/8.07 \\
    0.25 & 6.35/7.13 & 6.43/7.13 & 6.36/6.93 & 6.28/6.94 \\
    0.5 & 6.28/6.87 & 6.17/6.87 & 6.15/7.01 & 6.17/6.88 \\
    0.75 & 6.08/7.13 & 5.96/7.06 & 5.97/\textbf{6.71} & \textbf{5.88}/6.77 \\
    1 & 6.05/7.33 & 6.07/7.26 & 6.02/7.18 & 6.06/7.04 \\ \noalign{\hrule height 1.0pt}
    \end{tabular}
\end{table}

\unskip

\subsection{Ablation Study of DRNet}\label{sec4e}
\textls[-15]{We studied the impact of AD, SC and DIFF modules in the proposed DRNet architecture. }

To do this, we replaced AD with a new module called DRNet\_NoAD and used a parameter $\gamma$ to combine DIFF and SC outputs. We have formulated the new module using Equation (7) as follows:
\begin{equation}
{g_{DRNet\_NoAD}} = \gamma *{g_{sc}} + (1 - \gamma )*{g_{diff}}.
\end{equation} 
where $\gamma$ is a parameter to combine $g_{sc}$ and $g_{diff}$, $g_{sc}$ and $g_{diff}$is the output of SC and DIFF modules, $g_{DRNet\_NoAD}$ is the DRNet\_NoAD output.

DRNet\_NoAD used the loss function (i.e., Equation (5)) with $\alpha$ and $\beta$ values of 0.75. Table \ref{tab8}, reports the $angular-error$ of DRNet\_NoAD. It is noted that the value of $\gamma$ is automatically learned.
The results showed that the performance of DRNet in terms of $angular-error$ of 5.98 (MpiiGaze) and 6.71 (Eyediap) outperforms DRNet\_NoAD with 6.05 (MpiiGaze) and 7.16 (Eyediap). Therefore, the AD module has demonstrated a feasible impact on the DRNet model. 

\begin{table}
 \caption{The performance of DRNet\_NoAD.}\label{tab8}%
  \centering
  \begin{tabular}{lll}
    \toprule
     Dataset & Angular-Error & \boldmath{$\gamma$} \\ 
     \midrule
     MpiiGaze & 6.05 & 0.89 \\
     Eyediap & 7.16 & 0.88 \\
     \bottomrule
      \end{tabular}
\end{table}

\begin{table}
 \caption{The performance 
 of Diff-Nn in common environment.}\label{tab9}%
  \centering
  \begin{tabular}{lll}
    \toprule
        \textbf{Method} 
        & \textbf{ MpiiGaze (L/R/All)} 
        & \textbf{ Eyediap (L/R/All)} \\ 
        \midrule
        DRNet & 6.15/6.29/5.98 & 6.71/-/- \\
        Diff-Nn \cite{liu2019differential} & 10.73/10.92/10.83 & 11.82/-/-\\
        DRNet\_NoSC & 6.59/6.32/6.97 & 7.72/-/-\\
        DRNet\_NoAD & 6.11/6.25/6.05 & 7.16/-/-\\
        DRNet\_NoDIFF & 6.22/6.32/6.07 & 6.97/-/-\\
        \bottomrule
      \end{tabular}
\end{table}

A similar scenario was also considered for the SC module which was replaced by a new model, DRNet\_NoSC. 

We trained and tested DRNet\_NoSC with the left, right, and entire eyes. Here, it is noted that the Eyediap consists of the entire left images due to the preprocessing step. 
Using MpiiGaze/Eyediap datasets, DRNet achieves a better performance in terms of $angular-error$ of 5.98/6.71 compared to DRNet\_NoSC with 6.97/7.72 and Mnist model with 6.27/7.6. The result can be shown in Table ~\ref{tab9}.

We also studied the case of replacing the DIFF module by DRNet\_NoDIFF. The DRNet\_NoDIFF represents the gaze direction by summing AD and SC outputs. The results  in Table ~\ref{tab9} 
have shown that the DRNet architecture yields better performance in terms of $angular-error$ achieving 5.98/6.71 when compared to DRNet\_NoDIFF with 6.07/6.97.

It is worth noting that Diff-Nn \cite{liu2019differential} achieves the lowest performance. This is due to inference related to the selection of images. This suggests that the use of a residual structure such the proposed DRNet based auxiliary information is an attractive solution. Furthermore, directly predicting difference information is not a good choice in a common environment.

\section{Conclusions}\label{sec5}

This paper presents a novel appearance-based method (DRNet) architecture {that uses the shortcut connection to combine the original gaze direction and the difference information. A new loss function is proposed to evaluate the loss in 3D space}. 
DRNet outperforms the state-of-the-art in robustness to a noisy data set. { The experimental results demonstrate that DRNet can obtain the lowest $angular-error$ in MpiiGaze and Eyediap datasets by using eye features only, compared with the state-of-the-art gaze estimation methods. This paper provides a feasible solution to address the gaze calibration problem and enhance the robustness of noise images.} In future work, we will consider more factors in our DRNet model to improve the performance metrics, in particular, when we use facial and eye features. 

\section*{Acknowledgments}
This work was supported in part by the National Natural Science Foundation of China (61903090), Guangxi Natural Science Foundation (2022GXNSFBA035644) and the Foreign Young Talent Program (QN2021033002L).

\newpage
\bibliographystyle{unsrt}  
\bibliography{references}  

\end{document}